\documentclass[10pt,twocolumn,letterpaper]{article}

\usepackage{cvpr}
\usepackage{times}
\usepackage{epsfig}
\usepackage{graphicx}
\usepackage{amsmath}
\usepackage{amssymb}
\usepackage{xcolor}
\usepackage{subfigure}
 \usepackage{multirow}
\usepackage[pagebackref=true,breaklinks=true,letterpaper=true,colorlinks,bookmarks=false]{hyperref}

\cvprfinalcopy % *** Uncomment this line for the final submission

\def\cvprPaperID{9724} % *** Enter the CVPR Paper ID here

\ifcvprfinal\fi
\begin{document}

%%%%%%%%% TITLE
\title{Learning Depth from Monocular Videos Using Synthetic Data: A Temporally-Consistent Domain Adaptation Approach}

\author{\large Yipeng Mou$^1$   Mingming Gong$^{23}$   Huan Fu$^1$   Kayhan Batmanghelich$^2$   Kun Zhang$^3$   Dacheng Tao$^1$\\
$^1$UBTECH Sydney AI Centre, SIT, FEIT, The University of Sydney, Australia\\
$^2$Department of Biomedical Informatics, University of Pittsburgh\\
$^3$Department of Philosophy, Carnegie Mellon University\\
{\tt\footnotesize \{ymou5942@uni.,hufu6371@uni,dacheng.tao@\}sydney.edu.au \{mig73@, kayhan@\}pitt.edu kunz1@cmu.edu}
}

\maketitle
%\thispagestyle{empty}

%%%%%%%%% ABSTRACT
\begin{abstract}
   Majority of state-of-the-art monocular depth estimation methods are supervised learning approaches. The success of such approaches heavily depends on the high-quality depth labels which are expensive to obtain. Some recent methods try to learn depth networks by leveraging unsupervised cues from monocular videos which are easier to acquire but less reliable. In this paper, we propose to resolve this dilemma by transferring knowledge from synthetic videos with easily obtainable ground-truth depth labels. Due to the stylish difference between synthetic and real images, we propose a temporally-consistent domain adaptation (TCDA) approach that simultaneously explores labels in the synthetic domain and temporal constraints in the videos to improve style transfer and depth prediction. Furthermore, we make use of the ground-truth optical flow and pose information in the synthetic data to learn moving mask and pose prediction networks. The learned moving masks can filter out moving regions that produces erroneous temporal constraints and the estimated poses provide better initializations for estimating temporal constraints. Experimental results demonstrate the effectiveness of our method and comparable performance against state-of-the-art.
\end{abstract}

% !TEX root = egpaper_for_review.tex
%%%%%%%%% BODY TEXT
% monocular depth overview
% supervised, requires expensive labels, 
% unsupervised methods, stereo, monocular. 
% monocular problems, sensitive to moving objects, outlier.
% propose to use synthetic data, 1) domain mismatch image translation with frame consistent loss, supervised. 2) the depth estimation network supervised by grounth label on the translated image and ddvo on the real images. constrain each other 3) a moving mask predition to mask out the moving objects in videos to obtain further robustness.

\section{Introduction}
Monocular depth estimation is a fundamental problem in computer vision and 3d scene understanding. Appreciable progress has been made in recent years thanks to the deep convolutional neural networks(DCNNs) \cite{liu2016learning,wang2015towards,roy2016monocular,eigen2015predicting,kuznietsov2017semi,fu2018deep}. However, because most of these method consider depth estimation as a supervised learning problem, they require a large amount of images labeled with ground-truth depth maps, which are expensive to acquire in practice. To address the high cost issue, recent methods have investigated unsupervised approaches from stereo image pairs by recasting depth estimation as a reconstruction progress with the intermediate disparity prediction.   \cite{godard2017unsupervised,garg2016unsupervised,xie2016deep3d}.

% \begin{figure}[t]
% \begin{center}
%   \includegraphics[width=0.8\linewidth]{paper_image.jpg}
% \end{center}
%   \caption{Result ?? how to generate rgb depth?}
% \label{fig:long}
% \label{fig:onecol}
% \end{figure}

% However, recording depth ground truth is expensive: Current RGB-D datasets, such as NYU-depth\cite{Silberman:ECCV12}, KITTI\cite{Geiger2013IJRR}, and cityscapes\cite{cordts2016cityscapes}, are collected by depth sensors. As a result, the raw depth data are compromised due to the range limitation and low resolution, which introduces a large amount of post-processing.

Compared to stereo images, monocular videos are cheaper and even easier to obtain.  Recently, several unsupervised methods that trained solely on monocular video achieved promising performance. By incorporating  3d pose estimation and depth estimation, \cite{zhou2017unsupervised, wang2018learning} warp the image to neighbor frames and minimize the photometric consistency. However, this type of methods assume a rigid scene which brings two drawbacks to the model: 1) The moving region will generate incorrect projection and introduce noisy loss. 2) Due to the lack of motion information of moving objects, predicting the depth of moving region becomes a ill-posed problem and we have no clue to solve a unique depth value. Some fully unsupervised methods proposed to solve the moving problem by attaching a optical flow branch and deduct the moving mask \cite{luo2018every, ranjan2019competitive} . However, their methods only mask out the noisy loss while the moving object depth is still inaccurate without the help of stereo image. 

Inspired by \cite{li2019learning}, we hope to train the model with some direct supervision from single image rather than indirect clue like photometric loss. \cite{li2019learning} made a new dataset separating frozen people from crowds because building the synthetic dataset with moving human-being and camera motion is challenging. However, when it comes to the auto-driving setup, there is plenty of well-constructed synthetic dataset where cars and camera are moving naturally. Thus, instead of building a new dataset, we adopt domain adaptation methods for extra depth information.

The synthetic datasets perfectly solve the two previous problems: First, due to the nature of synthetic data, we can easily get a ground-truth moving mask so that the model has the ability to predict the mask and remove the noisy loss induced by moving regions. Second, the synthetic data provide pixel level depth ground-truth. Thus, the model will be trained under direct supervision and gain the ability to predict the depth of moving object.

There are some previous work applying domain adaptation to depth estimation \cite{atapour2018real, kundu2018adadepth, zheng2018t2net}. However,  those setups are all simply transferring the image or features ignoring the fact that geometry constraints can considerably improve both the style transfer and the task network. Many works has shown that the geometry constrains can significantly improve the quality of domain adaptation and task performance \cite{fu2018geometry}. In our work, we apply the geometry correspondence upon neighbor frames of transferred images and improve the quality of our performance.

In particular, to make more effective use of synthetic data, we propose a temporally-consistent domain adaptation (TCDA) approach that simultaneously explores labeled synthetic videos and monocular real videos. Our framework consists of a image translation (domain mapping) network, a depth prediction network, a moving mask prediction network, and a camera pose estimation network. The image translation network transforms the synthetic videos to real-style videos such that depth prediction network can be trained using ground-truth labels in the synthetic domain and temporal constraints in videos to reduce the structural distortion in the translation process. In addition, the moving mask prediction network uses the camera pose and optical flow information in the synthetic data. The predicted moving mask can be used to remove unreliable photometric losses for the moving pixels, which further purify the supervision information in real monocular videos. Finally, the camera pose estimation network trained on the ground-truth pose in the synthetic data can predict poses which provides better initializations for estimation of temporal constraints in real videos. The end-to-end training of image translation and depth estimation networks with the proposed losses improves the quality of translated images as well as depth estimation accuracy. We demonstrate the effectiveness of our method on the KITTI dataset~\cite{menze2015object} and the generalization performance on the Make3D dataset~\cite{saxena2009make3d}.

\begin{figure*}
\begin{center}
\includegraphics[width=0.9\linewidth]{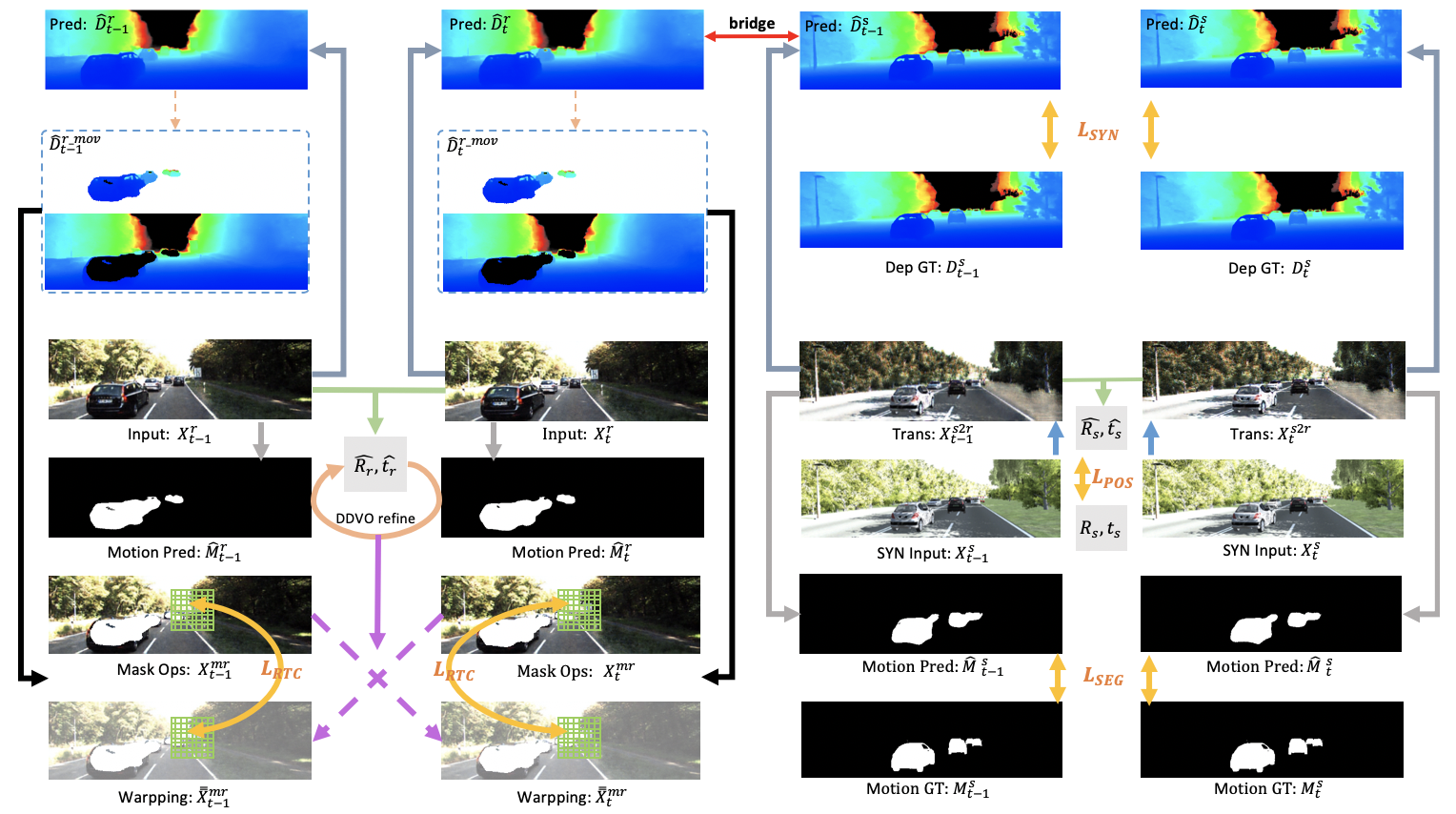}
\end{center}
   \caption{\label{fig:2} The proposed task network. Our model consists of three parts: a depth, a moving mask, and a relative camera pose predictor. They share the same feature from encoder E and have three different task decoder branches. $p_{pred}$ represents the relative pose from $x_{t}$ to $x_{t+1}$ predicted by our pose estimation network.}
   
\end{figure*}

\section{Related Work}

\paragraph{Monocular Depth Estimation}has been studied extensively over the past decade as it's important for understanding the 3D structure of scenes from 2D images. Early approaches relied on handcrafted features and incorporated global information by exploiting probabilistic graphical models ({\it e.g.}, MRFs) ~\cite{saxena2009make3d,saxena2006learning,liu2010single}, and nonparametric techniques~\cite{liu2014discrete,karsch2014depth,liu2011sift}. Thanks to the development of deep convolutional neural networks (DCNNs), recent methods developed various new network architectures for supervised monocular depth estimation~\cite{eigen2014depth,liu2016learning,he2018learning,xu2018structured,repala2018dual,qi2018geonet,cao2016estimating,laina2016deeper,roy2016monocular,chen2016single}. In the seminal work, Eigen {\it et al.}~\cite{eigen2014depth} developed the first depth estimation network that models multi-scale information. Up to now, there have been lots of follow-up works~\cite{liu2016learning,laina2016deeper,eigen2015predicting,li2015depth,xu2017multi,wang2015towards,fu2018deep} aiming at improving or extending this work in various directions. For example, the recent work \cite{fu2018deep} cast depth estimation as ordinal regression instead of regression and obtained state-of-the-art performance on several benchmarks.

An obvious disadvantage of supervised depth estimation is the requirement of large amounts of labeled images. To reduce the labeling cost, recent approaches sought for unsupervised methods from stereo image pairs or monocular videos \cite{xie2016deep3d,garg2016unsupervised,godard2017unsupervised,yin2018geonet,zhou2017unsupervised,wang2018learning}. Though these methods are termed as ``unsupervised" methods, they are different from unsupervised learning because the stereo pairs or monocular videos can provide weak supervision. In specific, Garg {\it et al.}~\cite{garg2016unsupervised} showed that unsupervised depth estimation can be supervised by a image reconstruction loss between stereo pairs. Following Garg {\it et al.}~\cite{garg2016unsupervised},  later works improved the way of supervision by exploiting left-right consistency~\cite{godard2017unsupervised}, semi-supervised learning ~\cite{kuznietsov2017semi}, etc. Regarding monocular videos, Zhou \etal \cite{zhou2017unsupervised} proposed a strategy that learns pose and depth CNN predictors by minimizing the photometric consistency between video frames during training. Wang \etal \cite{wang2018learning} further proposed a differentiable direct visual odometry (DDVO) approach that directly optimizes the pose and substantially boosted the depth prediction accuracy. 
%Some researchers also train monocular depth estimation together with other relevant tasks, like optical flow estimation, camera ego-motion and motion segmentation in a unsupervised manner. For example, the GeoNet network of Yin {\it et al.}~\cite{yin2018geonet} consists of three networks for depth estimation, camera motion estimation and optical flow estimation respectively, which can be learned jointly in an end-to-end manner. Ranjan {\it et al.}~\cite{ranjan2018adversarial} addressed such multi-task unsupervised learning in an adversarial collaboration.
%{\bf Semi-supervised Learning} attempts to make use of unlabelled data and labelled data jointly to improve the model performance. Kuznietsov {\it et al.}~\cite{kuznietsov2017semi} applied supervised learning to images with sparse ground truth depth as well as unsupervised learning to stereo pair images using the image reconstruction loss. Lai {\it et al.}~\cite{lai2017semi} proposed to train an optical flow estimation model on labelled synthetic data and unlabelled video data in a semi-supevised manner within a generative adversarial training framework. In this paper, we present a similar idea, but at the same time take into account the discrepancy between different datasets through integrating the domain adaptation procedure with the depth estimation task.
\paragraph{Domain Adaptation}aims to address the distribution shift issue such that model trained on a dataset can be generalized to a different but related dataset \cite{pan2010survey}. A large body of recent works tried to learn a domain-invariant representation via DCNNs ~\cite{ganin2015unsupervised,ganin2016domain,icml2015_long15,ajakan2014domain,sun2016deep}. These methods rely on various distance discrepancy measures as objective functions to match representation distributions; typical ones include maximum mean discrepancy (MMD) \cite{icml2015_long15}, separability measured by classifiers \cite{ganin2016domain}, and optimal transport \cite{courty2017optimal,damodaran2018deepjdot}. 

%Recent success of GANs~\cite{goodfellow2014generative} in image generation~\cite{isola2017image,zhang2018self,xu2018attngan,liu2016coupled,CycleGAN2017} has motivated researchers to exploit adversarial learning for domain adaptation~\cite{Hoffman_cycada2017,tsai2018learning,chen2018domain,hong2018conditional}.  Hoffman {\it et al.}~\cite{Hoffman_cycada2017} proposed pixel-level and feature-level domain adaptation for semantic segmentation. Pixel-level adaptation, based on CycleGAN~\cite{CycleGAN2017}, translates source data ({\it i.e.} synthetic data) into target data ({\it i.e.} real data), while feature-level adaptation is similar to the approaches learning domain-invariant feature representations aforementioned, but is improved by using an adversarial loss. To overcome the semantic inconsistency issue introduced by the process of adaptation in the pixel space, the authors defined a semantic consistency constraint to enforce pixel-level adaptation to preserve semantic information, which is crucial for semantic segmentation.
Coming to DA for depth estimation, Atapour {\it et al.}~\cite{atapour2018real} developed a two-stage method which first learned a image translator \cite{zhu2017unpaired} to stylize the real images into synthetic images, and then trained a supervised depth estimation network using the original synthetic images. Kundu {\it et al.}~\cite{kundu2018adadepth} proposed a content congruent regularization method to address the model collapse problem which usually happens in high-dimensional data. Recently, Zheng {\it et al.}~\cite{zheng2018t2net} developed an end-to-end adaptation network, {\it i.e.} ${\rm T^2Net}$,  where the translation network and the depth estimation network are optimized jointly so that they can improve each other. However, these works overlooked the temporal constraints from both synthetic and real domain monocular videos , thus produced unsatisfactory image translation quality. 
\section{Proposed Method}
In this section, we will first take a brief review of how to estimate depth from monocular videos. Then we further claim our motivation based on an observation from a small experiment. Finally we will present the proposed depth estimation framework in detail.

\subsection{Motivation}\label{sec:moti}
The goal here is to recover depth from monocular videos in an unsupervised fashion. A straightforward idea is to recast depth estimation as a reconstruction problem by modeling the temporal consistency across neighbored frames. Mathematically, let $p_t$ denotes the positions of a single pixel of frame t, the position of the corresponding pixel in t+1 can be computed as $p_{t+1} = KT_{(t,t+1)}D_t(p_t)K^{-1}p_t$, where K and T represent the camera poses, D is the depth map. Thus, we can give indirect supervisions to both a camera pose network and a depth estimation network iteratively via a photometric loss on warping image.

As suggested by the above literature review, solid DE from MVs relies on accurate camera pose estimation and the robustness of the photometric constraint. Unluckily, practical dynamic scenes may contain many objects with irregular movement, which would significantly degrade the performance for both of them. In fact, even with relatively reliable camera pose \cite{wang2018learning}, it is not authentic to directly enforce photometric consistency on moving regions. To alleviate this issue, several works \cite{luo2018every, ranjan2019competitive} exploited the  to predict non-rigid and rigid moving (displacement caused by camera moving respectively), and (softly) mask out real moving regions when computing photometric loss. However, while removing noises (real moving objects) would naturally achieve more accurate depth estimation on rigid regions, these approaches do not perform well on non-rigid regions due to lack of extra supervisions on these removed moving regions. 
To provide a clearer illustration, we examined our observation on ApolloScape dataset \cite{wang2019apolloscape} via an experiment. Taking the algorithm presented in \cite{zhou2017unsupervised} as an example, we train two models with one optimized by masked (using ground truth moving masks) photometric and SSIM losses (Robust-SfMLearner), and the other one the vanilla version (SfMLearner). Specifically, from the reported scores in Table \ref{tab:small_exp}, both the consistent improvement on rigid regions and the degraded performance on non-rigid regions further support our analysis before. Importantly, we observe that lots of predicted depth values for moving cars is extremely large in both versions of the experiments. One of the main reasons is that vehicles have similar moving trend in both direction and speed as the cameras, which results in small disparity (i.e., large depth) estimation between neighbored frames without extra supervisions on these real moving regions. 
Motivated by the observations, we proposed to exploit indirect supervisions for moving regions and improve the robustness of photometric loss for depth estimation by introducing synthetic benchmarks and employing domain adaptation techniques. In the following, we will first present the overall architecture of our novel approach to unsupervised depth estimation from monocular videos and then explain how to take fully advantage of synthetic dataset to provide a solution for the issues aforementioned.

\begin{table}[]\centering
\begin{tabular}{|l|c|c|}
\hline
\multirow{2}{*}{} & \multicolumn{2}{c|}{RMSE} \\ \cline{2-3} 
                  & Robust-SfMLearner                  & SfMLearner \cite{zhou2017unsupervised}                  \\ \hline
Static        & 7.3158                  & 5.31                    \\ \hline
Moving        & 13.0642                 & 12.5865                 \\ \hline
All Area          & 8.7741                  & 7.1775                  \\ \hline
\end{tabular}
\bigskip
\caption{An illustration experiment on ApolloScape dataset.}
\label{tab:small_exp}
\end{table}

\subsection{Overall Network Architecture}
The overall network architecture, as shown in Figure 3, mainly consists of two sub-networks: 1) A temporal consistency guided image-to-image translation network which maps the images in the synthetic domain to images that have the same style as target domain, and 2) a depth prediction network that is driven by ground truth labels in  synthetic domain and robust temporal constraints building on the real-domain videos. Note that, the intermediate predictions from source domain including moving masks and camera pose coupled with the domain adaptation techniques are significant to formulate reliable unsupervised constraints for target domains as analyzed before. Here, we will clearly introduce the two parts, respectively. More implementation details are provided in the supplemental materials.

\begin{figure}
\begin{center}
\includegraphics[width=1\linewidth]{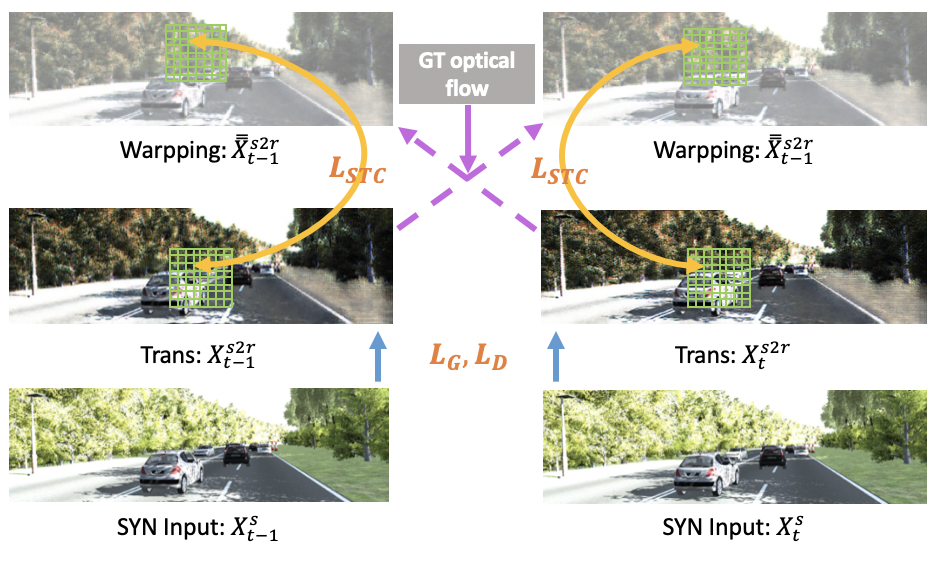}
\end{center}
   \caption{\label{fig:2} An illustration of the temporal consistency guided I2I translation network. Please refer Sec.~\ref{sec:i2i} for details.}
\end{figure}

\subsubsection{Temporal Consistency Guided I2I Translation}\label{sec:i2i}
To make better use of the synthetic data for depth estimation, an essential issue is how to reduce the discrepancy between synthetic domain and real domain. Previous works \cite{atapour2018real} prefer to employ CycleGAN \cite{CycleGAN2017}, an  unsupervised image-to-image translation approach, to provide a partial solution by translating synthetic images to having the same style as real images while preserving the semantic structures. However, one of the main disadvantages is that learning CycleGAN is memory intensive due to the two-directional translation strategies, which makes it not possible to train a deep domain adaptation dense prediction network in an end-to-end fashion \cite{Hoffman_cycada2017}. Instead, we suggest that learning an one-sided mapping by enforcing a simple but effective optical flow guided temporal consistency loss is sufficient to generate reasonable translations in our task. In fact, the supervised depth prediction branch in the synthetic domain could also prevent mode collapse and semantic distortions in a way.

Mathematically, given two successive frames $x^s_{t}$ and $x^s_{t+1}$ in the synthetic video, the translation network $G$ maps the synthetic frames to $G(x^s_{t})$ and $G(x^s_{t+1})$. Our I2I translation networks are optimized by three objectives.

\paragraph{Adversarial constraint} The first item is the standard adversarial loss used in GAN:
 \begin{flalign}\label{Eq:gan_loss}
 L_{GAN}(G,D)=&{\mathbb{E}}_{x^r\sim P_R(x^r)} [\log( D(x^r))]\nonumber\\&+{\mathbb{E}}_ {x^s\sim P_S(x^s)}[\log(1- D(G(x^s)))],
 \end{flalign}
 where $D$ is a discriminator network, $x^r$ denotes real images, and $P_R$ and $P_S$ represent the probability distributions of real and synthetic images, respectively. Here, we consider each frame independently by ignoring the temporal relations.
 
 \paragraph{Flow Guided Synthetic Temporal Consistency} The second objective is the temporal consistency loss designed according to the geometric relations between the two input synthetic frames:
 \begin{flalign}\label{Eq: syn_temporal}
& L_{STC}(G) =\nonumber\\&{\mathbb{E}}_{x_t^s,x_{t+1}^s\sim P_S(x_t^s,x_{t+1}^s)}[\|F(G(x^s_{t}))-G(x^s_{t+1})\|_1],
 \end{flalign}
where $F(\cdot)$ is a function warping an image based on the optical flow. This loss could preserve that geometric relations between the translated images.

\paragraph{Identity constraint} The last constraint is a widely used identity loss that stabilizes the translation process \cite{zheng2018t2net,CycleGAN2017}. The identity loss is defined as
\begin{equation}\label{Eq: identity}
L_I = {\mathbb{E}}_{x^r\sim P_R(x^r)}[\|G(x^r)-x^r\|_1],
\end{equation}
which forces the translator $G$ to be an identity mapping w.r.t. real domain images.
% Our approach introduces synthetic data to improve the SFM depth prediction. At the same time, two neighbor frames of synthetic images and two neighbor frames of real images are applied. Inherited from T2net\cite{zhou2017unsupervised}, the twin pipeline structure are applied for both frames of synthetic and real images. (Shown in figure 2). The whole approach contains three parts:

\subsubsection{Robust Depth Prediction} \label{sec:rob_dep_pred}

Our depth prediction network follows a standard encoder-decoder structure as previously \cite{ronneberger2015u}. The network parameters are jointly optimized by a supervised L1 loss with image-depth pairs provided by synthetic domain, and an unsupervised robust photometric loss according to adapted camera pose and moving masks in real domain.

We indicate that the synthetic domain provides various easily captured ground truth labels (e.g., depth, segmentation, optical flow, and camera pose) but low quality translated images, while the real domain contains high quality images but unreliable photometric losses due to moving objects in the scene. Thus, as analyzed in Sec. \ref{sec:moti}, exploiting the complementary between each other reasonably would definitely improve the depth prediction accuracy. We will detail the jointly learning strategy of our novel robust depth prediction network.

\paragraph{Depth Regression} Optimizing the depth prediction model in synthetic domain is straightforward. Since we are offered ground truth depth labels for each image,  we directly employ L1 loss to measure the depth prediction error:
\begin{flalign}\label{Eq: syn_depth_loss}
&L_{SYN}(E,D_d,G) = \nonumber\\ & {\mathbb{E}}_{(x^s,d^s_{gt})\sim P_S(x^s,d^s_{gt})}[\|D_d(E(G(x^s)))-d^s_{gt}\|_1],
\end{flalign}
where $E$ represents the encoder which is shared by all following sub-tasks, $D_d$ denotes the depth decoder. Note that, this branch can give direct supervision for moving regions, thus provide a remedy to the issue coming from masked photometric loss.

\paragraph{Camera Pose Prediction}
As explained in Sec.~\ref{sec:moti}, the temporal photometric loss is sensitive to the estimation of camera poses. To provide a better initialization, we propose to learn a pose prediction network from the translated synthetic data. In specific, we take the features extracted by the encoder $E$ from translated images as inputs and map the features to pose parameters $\mathbf{p}$ using a pose prediction network $D_p$. The pose prediction network is driven by a standard $L_1$ loss. Note that, We also employ the differentiable direct visual odometry (DDVO) method \cite{wang2018learning} to refine the estimated camera pose during the training procedure. 

\paragraph{Moving detection}
In addition, we need to perform moving detection to formulate a moving robust photometric loss. Previous works prefer to estimate optical flow to model irregular movements \cite{luo2018every, ranjan2019competitive}. However, on the one hand, obtaining dense flow annotations for real videos is as expensive as capturing depth annotations. On the other hand, there is still a large gap between supervised and unsupervised optical flow estimation. Here, we propose a potential solution by building a bridge between synthetic domain and real domain. In specific, we first recast moving detection as a segmentation task by generating moving mask ground truth according to the provided ground truth labels in synthetic data, i.e., optical flow, instance segmentation, camera extrinsic parameters \footnote{It is convenient for synthetic data to offer various ground truth labels.}. Then, we adopt the aforementioned domain adaptation technique to handle moving in real videos. As a result, the loss function for the moving detection branch is expressed as:
\begin{flalign}\label{Eq: SEG_LOSS}
&L_{SEG}(M) = \nonumber\\&{\mathbb{E}}_{(x^s,m_{gt}^s)\sim P_S(x^s,m_{gt}^s)}[\|m_{gt}*\log(M(G(x^s)))\|_1]
\end{flalign}

%\begin{flalign}\label{Eq: syn_depth_loss}
%&L_{SYN}(E,D_d,G) \nonumber\\ &= %{\mathbb{E}}_{(x^s,d^s_{gt})\sim %P_S(x^s,d^s_{gt})}[\|D_d(E(G(x^s)))-d^s_{gt}\|_1],
%\end{flalign}
% Specifically, given the outputs from the depth prediction network and the relative camera pose $\mathbf{p}_t$, we have the following loss:
% \begin{flalign}\label{Eq:photo_loss}
% L_{PH}(\mathbf{p}_t,D_d,E_d,G) &= \mathbb{E}_{x_t^r,x_{t+1}^r\sim P_S(x_t^r,x_{t+1}^r)}\nonumber\\ &\|\mathcal{W}(x^r_{t}, \mathbf{p}, D_d(E_d(G(x_t^r))))-x^s_{t+1}\|^2,
% \end{flalign}
\paragraph{Temporal Photometric Consistency}  Given the successive frames $x^r_{t}$ and $x^r_{t+1}$ from monocular videos, we perform indirect supervision to the depth prediction network by modeling temporal photometric consistency as previously \cite{wang2018learning}. However, the standard photometric loss in DVO is sensitive to moving objects which violates the temporal consistency as analysed before. To provide a partial remedy, we propose to consider the moving prior when computing the photometric loss. In other words, we predict moving masks via DA techniques, and develop a robust photometric constraint by ignoring the moving objects. Note that, we have extra supervisions from synthetic domain for these moving regions, thereby the issue (unreliable depth estimation for moving regions) caused by the robust photometric loss can also be addressed (refer to Sec.~\ref{sec:moti}).
Finnaly, given the predicted depth $D_d$, the relative camera pose $\mathbf{p}_{t}^r$, and estimated moving masks $M_{pred,t}^r$ and $M_{pred,t+1}^r$ for frames $t$-th and $t+1$-th, the robust temporal consistency loss is defined as:
\begin{flalign}\label{Eq:photo_loss}
&L_{RTC}(\mathbf{p}_{t}^r,D_d,E,G) = \mathbb{E}_{x_t^r,x_{t+1}^r\sim P_R(x_t^r,x_{t+1}^r)}\nonumber\\ &\|\mathcal{W}(M^{r}_{pred,t}\circ x^r_{t}, \mathbf{p}_t^r, D_d(E(G(x_t^r)))-M^{r}_{pred,t+1}\circ x^r_{t+1})\|^2,
\end{flalign}
where $\circ$ denotes element-wise product, $\mathcal{W}$ is a warping function that maps $x_t$ to $x_{t+1}$ according to the estimated depth and camera pose. 
\begin{table*}[t]\small
\begin{center}
\begin{tabular}{|c|c|c|cccc|ccc|}
\hline
\multirow{2}{*}{Method} &
\multirow{2}{*}{Dataset} &
\multirow{2}{*}{Supervised} &
\multicolumn{4}{c|}{Error Metrics}&
\multicolumn{3}{c|}{Accuracy Metrics}\\ \cline{4-10} &
&
& 
Abs Rel&
Sq rel&
RMSE&
RMSE log &
$\sigma < 1.25 $ &
$\sigma < 1.25^2 $ &
$\sigma < 1.25^3 $                    \\ \hline
\multicolumn{10}{|c|}{depth capped at 80m}\\
\hline
\multicolumn{1}{|c|}{Eigen \emph{et al.} \cite{eigen2015predicting}}&
K&
Yes&
\multicolumn{1}{c}{0.203}&
\multicolumn{1}{c}{1.548}&
\multicolumn{1}{c}{6.307}&
\multicolumn{1}{c|}{0.282}&
\multicolumn{1}{c}{0.702}&
\multicolumn{1}{c}{0.890}&
\multicolumn{1}{c|}{0.958} \\ \hline
\multicolumn{1}{|c|}{Godard \emph{et al.} \cite{godard2017unsupervised}}&
K&
No&
\multicolumn{1}{c}{0.148}&
\multicolumn{1}{c}{1.344}&
\multicolumn{1}{c}{5.927}&
\multicolumn{1}{c|}{0.247}&
\multicolumn{1}{c}{0.803}&
\multicolumn{1}{c}{0.922}&
\multicolumn{1}{c|}{0.964} \\ \hline

\multicolumn{1}{|c|}{DDVO \cite{wang2018learning}}&
K&
No&
\multicolumn{1}{c}{0.151}&
\multicolumn{1}{c}{1.257}&
\multicolumn{1}{c}{5.583}&
\multicolumn{1}{c|}{0.228}&
\multicolumn{1}{c}{0.810}&
\multicolumn{1}{c}{0.936}&
\multicolumn{1}{c|}{0.974} \\ \hline
\multicolumn{1}{|c|}{Zhou \emph{et al.} \cite{zhou2017unsupervised}}&
K&
No&
\multicolumn{1}{c}{0.208}&
\multicolumn{1}{c}{1.768}&
\multicolumn{1}{c}{6.856}&
\multicolumn{1}{c|}{0.283}&
\multicolumn{1}{c}{0.678}&
\multicolumn{1}{c}{0.885}&
\multicolumn{1}{c|}{0.957} \\ \hline

\multicolumn{1}{|c|}{Kundu \emph{et al.} \cite{nath2018adadepth}}&
K+V&
No&
\multicolumn{1}{c}{0.214}&
\multicolumn{1}{c}{1.932}&
\multicolumn{1}{c}{7.157}&
\multicolumn{1}{c|}{0.295}&
\multicolumn{1}{c}{0.665}&
\multicolumn{1}{c}{0.882}&
\multicolumn{1}{c|}{0.950} \\ \hline

\multicolumn{1}{|c|}{Kundu \emph{et al.} \cite{nath2018adadepth}}&
K+V&
semi&
\multicolumn{1}{c}{0.167}&
\multicolumn{1}{c}{1.257}&
\multicolumn{1}{c}{5.578}&
\multicolumn{1}{c|}{0.237}&
\multicolumn{1}{c}{0.771}&
\multicolumn{1}{c}{0.922}&
\multicolumn{1}{c|}{0.971} \\ \hline

\multicolumn{1}{|c|}{Zheng \emph{et al.} \cite{zheng2018t2net}}&
K+V&
No&
\multicolumn{1}{c}{0.174}&
\multicolumn{1}{c}{1.410}&
\multicolumn{1}{c}{6.046}&
\multicolumn{1}{c|}{0.253}&
\multicolumn{1}{c}{0.754}&
\multicolumn{1}{c}{0.916}&
\multicolumn{1}{c|}{0.966} \\ \hline
\multicolumn{1}{|c|}{Sensitive-TCDA}&
K+V&
No&
\multicolumn{1}{c}{0.155}&
\multicolumn{1}{c}{1.144}&
\multicolumn{1}{c}{5.578}&
\multicolumn{1}{c|}{0.229}&
\multicolumn{1}{c}{0.794}&
\multicolumn{1}{c}{0.931}&
\multicolumn{1}{c|}{0.974} \\ \hline
\multicolumn{1}{|c|}{TCDA}&
K+V&
No&
\multicolumn{1}{c}{\textbf{\textcolor{blue}{0.145}}}&
\multicolumn{1}{c}{\textbf{\textcolor{blue}{1.058}}}&
\multicolumn{1}{c}{\textbf{\textcolor{blue}{5.291}}}&
\multicolumn{1}{c|}{\textbf{\textcolor{blue}{0.215}}}&
\multicolumn{1}{c}{\textbf{\textcolor{blue}{0.816}}}&
\multicolumn{1}{c}{\textbf{\textcolor{blue}{0.941}}}&
\multicolumn{1}{c|}{\textbf{\textcolor{blue}{0.977}}} \\\hline
\multicolumn{10}{|c|}{depth capped at 50m}\\

\hline

\multicolumn{1}{|c|}{Garg\cite{garg2016unsupervised}}&
K&
No&
\multicolumn{1}{c}{0.169}&
\multicolumn{1}{c}{1.080}&
\multicolumn{1}{c}{5.104}&
\multicolumn{1}{c|}{0.273}&
\multicolumn{1}{c}{0.740}&
\multicolumn{1}{c}{0.904}&
\multicolumn{1}{c|}{0.962} \\ \hline

\multicolumn{1}{|c|}{Kundu \emph{et al.} \cite{nath2018adadepth}}&
K+V&
No&
\multicolumn{1}{c}{0.203}&
\multicolumn{1}{c}{1.734}&
\multicolumn{1}{c}{6.251}&
\multicolumn{1}{c|}{0.284}&
\multicolumn{1}{c}{0.687}&
\multicolumn{1}{c}{0.899}&
\multicolumn{1}{c|}{0.958} \\ \hline

\multicolumn{1}{|c|}{Kundu \emph{et al.} \cite{nath2018adadepth}}&
K+V&
semi&
\multicolumn{1}{c}{0.162}&
\multicolumn{1}{c}{1.041}&
\multicolumn{1}{c}{4.344}&
\multicolumn{1}{c|}{0.225}&
\multicolumn{1}{c}{0.784}&
\multicolumn{1}{c}{0.930}&
\multicolumn{1}{c|}{0.974} \\ \hline

\multicolumn{1}{|c|}{Zheng \emph{et al.} \cite{zheng2018t2net}}&
K+V&
No&
\multicolumn{1}{c}{0.168}&
\multicolumn{1}{c}{1.199}&
\multicolumn{1}{c}{4.674}&
\multicolumn{1}{c|}{0.243}&
\multicolumn{1}{c}{0.772}&
\multicolumn{1}{c}{0.912}&
\multicolumn{1}{c|}{0.966} \\ \hline

\multicolumn{1}{|c|}{Sensitive-TCDA}&
K+V&
No&
\multicolumn{1}{c}{0.149}&
\multicolumn{1}{c}{0.879}&
\multicolumn{1}{c}{4.191}&
\multicolumn{1}{c|}{0.216}&
\multicolumn{1}{c}{0.806}&
\multicolumn{1}{c}{0.940}&
\multicolumn{1}{c|}{0.978} \\ \hline

\multicolumn{1}{|c|}{TCDA}&
K+V&
No&
\multicolumn{1}{c}{\textbf{0.139}}&
\multicolumn{1}{c}{\textbf{0.814}}&
\multicolumn{1}{c}{\textbf{3.995}}&
\multicolumn{1}{c|}{\textbf{0.203}}&
\multicolumn{1}{c}{\textbf{0.830}}&
\multicolumn{1}{c}{\textbf{0.949}}&
\multicolumn{1}{c|}{\textbf{0.980}} \\ \hline
\end{tabular}
\end{center}
\caption{The result is evaluated on the Eigen \emph{et al.} \cite{eigen2015predicting} split of KITTI\cite{Geiger2013IJRR}. Here, K represents KITTI, V is vKITTI (sythetic dataset), and Sensitive-TCDA denotes our model with the standard photometric constraint.}
\label{tab:kitti}
\end{table*}

\begin{center}
\begin{table}[t]\footnotesize
\centering
\begin{tabular}{|c|c|cccc|}
\hline
\multirow{2}{*}{Method} & \multirow{2}{*}{Train} & \multicolumn{4}{c|}{Error Metrics (the lower the better)} \\ \cline{3-6} &
& Abs Rel       &
Sq Rel      &
RMSE       &
RMSE(log)      \\ \hline
Mean          &
NA         &
0.876         &
13.98       &
12.27      &
0.307          \\ \hline
Karsch \emph{et al.} \cite{karsch2014depth}          &
Yes         &
0.428         &
5.079       &
8.389      &
0.149          \\ \hline
Laina \emph{et al.} \cite{laina2016deeper}          &
Yes         &
0.204         &
1.840       &
5.683      &
0.084          \\ \hline
Godard \emph{et al.} \cite{godard2017unsupervised}          &
No         &
0.544         &
10.94       &
11.76      &
0.193          \\ \hline
Zhou \emph{et al.} \cite{zhou2017unsupervised}          &
No         &
\textbf{0.383}         &
5.321       &
10.47      &
0.478          \\ \hline
DDVO \cite{wang2018learning}          &
No         &
0.387         &
4.720       &
8.09      &
0.204          \\ \hline
Zheng \emph{et al.} \cite{zheng2018t2net}          &
No         &
0.428         &
5.132       &
8.926      &
0.208          \\ \hline
TCDA          &
No         &
0.384         &
\textbf{3.885}       &
\textbf{7.645}      &
\textbf{0.181}          \\ \hline
\end{tabular}
\bigskip
\caption{Performance on Make3D. The ``train" column states whether the method is trained on make3d training set. Metrics is computed in the central image crop with depth capped at 70m.}
\label{tab:make3d}
\end{table}
\end{center}
% !TEX root = egpaper_for_review.tex

\section{Experiments}
In this section, we first present the implementation details including the network architectures, the data-preprocessing methods, and the training/inference strategies. Then we demonstrate the effectiveness of our model on two well-known and challenging benchmarks (i.e., KITTI\cite{Geiger2013IJRR} and Make3D~\cite{saxena2006learning,saxena2009make3d}) by making a comparison with the baselines and previous state-of-the-art approaches.  Finally, we perform various ablation experiments to comprehensively study the behavior of our method.
\subsection{Implementation Details}\label{sec:implement}
\paragraph{Network Architecture}
The proposed model mainly consists of a style transfer module and a depth prediction module. For the style transfer module, we used the network architectures from T2Net \cite{zheng2018t2net} and remove the feature GAN part. The depth prediction module is composed of three main sub-networks including a depth estimation network, a moving detection network, and a camera pose prediction network. The structures of the former two follow UNet~\cite{ronneberger2015u} fashion, and the development of the camera pose network is inspired by \cite{zhou2017unsupervised}. 

\paragraph{Data Pre-processing}
We perform the proposed TCDA method by employing vKITTI (labeled) as the source domain and KITTI (unlabeled) as the target domain. We collect 40000 and 12000 pairs of neighbor frames from KITTI and vKITTI, respectively. The images are then resized to 192$\times$640 in both our training and inference stages.

In our setting, while several kinds of ground truth labels (i.e., moving mask, camera pose, and depth) are accessible for vKITTI images, KITTI can only provide some video sequences. We normalize the ground truth depth from [0, 80] to [-1, 1]. The pixels with depth greater than 80m are labeled as 1 in our training stage.

\begin{figure*}
\begin{center}
\includegraphics[width=0.85\linewidth]{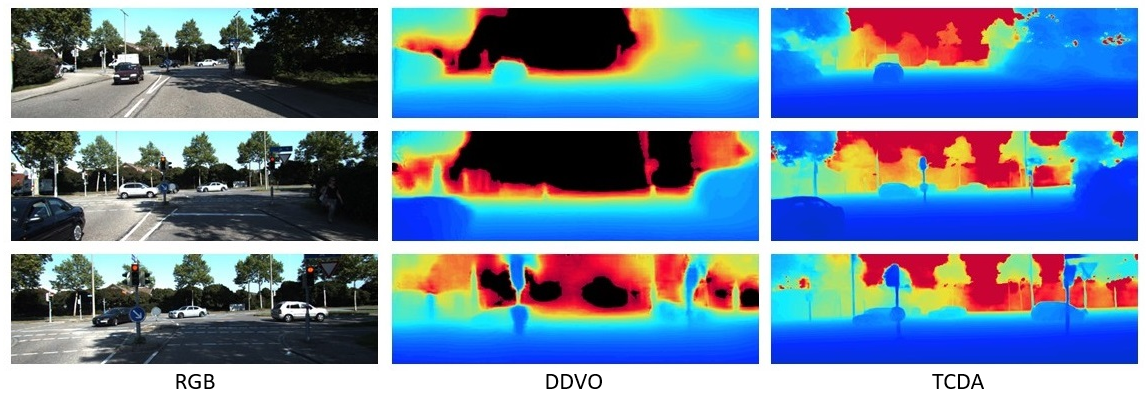}
\end{center}
   \caption{Dynamic scene outcome: The qualitative comparison between DDVO \cite{wang2018learning} and our TCDA.}
\label{fig:kitti_output}
\end{figure*}

\begin{figure}
\begin{center}
\includegraphics[width=1\linewidth]{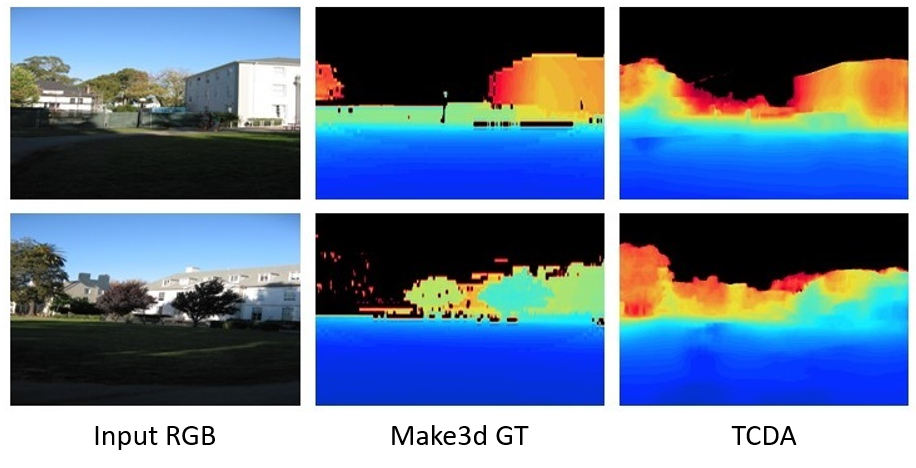}
\end{center}
   \caption{Qualitative Reuslt on Make3D dataset. The groundtruth is interpolated for demonstration. }
\label{fig:make3d_output}
\end{figure}

\paragraph{Training strategy}
As aforementioned, our model deeply relies on generative adversarial networks. However, the training of GANs is currently known to be unstable. Thus, we cannot directly optimize the whole framework in an end-to-end fashion from scratch. Instead, we pre-train the naive depth adaptation model which only contains a style transfer network and a depth adaptation network (i.e., the left part in Figure \ref{fig:2}). Then, we train the camera pose network and the moving detection network for some epochs using the translated images by freezing the parameters that were obtained in the previous stage. Finally, we introduce the DDVO block and fine-tune the whole framework in an end-to-end manner. Note that, the pre-training stages can offer a relatively robust initialization to prevent the model from generating unreasonable translations and depth predictions caused by the instability of GANs.

%For the pre-training stages, we apply a learning rate of 1e-4 for the depth network and a learning rate of 2e-5 for the translation network for the first 16 epochs and then performed cosine annealing for the next 16 epochs. Then we fix the translation network and train the depth network for 32 epochs with learning rate 1e-5 applying cosine annealing and warm restart at 1, 2, 4, 8, 16 epochs. Finally, we finetune the network for 8 epochs with learning rate of 1e-6. We adopt the Adam optimizer in our implementation. 

\subsection{Benchmark performance: KITTI}
As done previously, we evaluate our method on the KITTI Eigen split (697 images) at the distances of 80m and 50m, respectively. The evaluation codes and the center cropping strategies are provided by \cite{zhou2017unsupervised}. 
The scores are reported in Table \ref{tab:kitti}. Our method outperforms previous UDA approaches \cite{zheng2018t2net,atapour2018real,kundu2018adadepth} and previous monocular video approaches \cite{zhou2017unsupervised,wang2018learning} by a convincing margin. Importantly, our model yields higher scores than previous state-of-the-art DDVO \cite{wang2018learning} (video depth estimation algorithm) due to our two aspects of refinement of photometric loss. We also make qualitative comparisons with DDVO \cite{wang2018learning}. As shown in Figure \ref{fig:kitti_output}, DDVO cannot identify the moving cars around the traffic light. In contrast, our model can recognize most of the moving vehicles. In addition, the outlines of objects are preserved in our method, while DDVO often produces blurry outputs.

\subsection{Benchmark performance: Make3D}
We study the generalization capability of our model on the Make3D \cite{saxena2006learning,saxena2009make3d} test dataset. The scores are reported in Table \ref{tab:make3d}. Despite the large domain shift between KITTI and Make3D, our method can still produce reasonable predictions, and generally performs better than previous state-of-the-art methods. Note that, the evaluated model is not trained or fine-tuned using the Make3D images. Some qualitative results are shown in Figure \ref{fig:make3d_output}.

\begin{table*}[]\centering\small
\begin{tabular}{|l|l|c|c|c|c|c|c|c|}
\hline
\multirow{2}{*}{}          & \multirow{2}{*}{Method} & \multicolumn{4}{c|}{Error metric}   & \multicolumn{3}{c|}{Accuracy metric} \\ \cline{3-9} 
                           &                         & Abs Rel & Sq Rel & RMSE  & RMSE log & $\sigma < 1.25 $         & $\sigma < 1.25^2 $         & $\sigma < 1.25^3 $         \\ \hline
\multirow{2}{*}{SFM}  & SFM                     & 0.272       & 1.793      & 5.048     & 0.288        & 0.652          & 0.888          & 0.957          \\ \cline{2-9} 
                           & Robust-SFM                & 0.165       & 1.004      & 4.795     & 0.231        & 0.770          & 0.929          & 0.975           \\ \hline
\multirow{5}{*}{DA} & NO\_DA                   & 0.284   & 2.459  & 6.816 & 0.382    & 0.508      & 0.780      & 0.907      \\ \cline{2-9} 
                           & UDA                     & 0.172   & 1.191  & 4.741 & 0.250    & 0.765      & 0.910      & 0.965      \\ \cline{2-9} 
                           & DA-RTC                  & 0.161   & 0.997  & 4.208 & 0.223    & 0.798      & 0.932      & 0.974      \\ \cline{2-9} 
                           & Sensitive-TCDA                 & 0.149   & 0.879  & 4.191 & 0.216    & 0.806      & 0.940      & 0.978      \\ \cline{2-9} 
                           & TCDA (DA-RTC-STC)        & 0.139   & 0.814  & 3.995 & 0.203    & 0.830      & 0.949      & 0.980      \\ \hline
\end{tabular}
\bigskip
\caption{The results of different ablation study we conducted. The performance is evaluated on KITTI Eigen split and the depth is capped at 50m. STC stands for synthetic temporal consistency, RTC denotes real temporal consistency. We first compare Robust-SFM and SFM where the photometric constraint are refined or not refined by moving masks. Then we introduce the DA methods: NO\_DA represents model without translation network, UDA is vanilla domain adaptaion method. DA-RTC stands for only applying the temporal consistency upon real domain. Sensitive-TCDA is TCDA trained with standard photometric loss.}
\label{tab:ablation}
\end{table*}

\section{Ablation Study}
To further analyze our method, we conduct several ablation studies to discuss each part of our structure and demonstrate their outcomes. We evaluate these variants on the KITTI test set and report the scores in Table \ref{tab:ablation}.

\subsection{Robust Photometric Constraint}
We demonstrate that standard photometric loss is sensitive to moving regions. We take a conventional SFM based depth estimation network \cite{zhou2017unsupervised} as an example, and perform SFM and Robust-SFM respectively. Here, Robust-SFM represnts SFM driven by the robust photometric loss, where the moving regions are masked out. We pre-compute the moving masks according to the pre-trained moving detector network of TCDA. From the scores reported in Table \ref{tab:ablation}, moving guided photometric loss can significantly improve SFM by a large margin

\subsection{Domain Adaptation}
We then study the importance of domain adaptation by making a comparison between No\_DA and UDA Here, No\_DA represents the model that is directly trained via synthetic images without adaptation and corresponding ground truth depth maps provided by vKITTI. UDA is the naive domain adaptation model which is optimized the GAN loss (\ref{Eq:gan_loss}), the Identity loss (\ref{Eq: identity}), and the synthetic depth loss(\ref{Eq: syn_depth_loss}). As shown in Table \ref{tab:ablation}, No\_DA performs badly on real images (target domain) due to the large domain shift between vKITTI and KITTI, while UDA provides an effective remedy to this issue.

\subsection{Temporal Consistency}
We investigate the effectiveness of temporal consistency losses by examining TCDA together with two variants including UDA and DA-RTC. In specific, we formulate DA-RTC by integrating the real temporal consistency loss (\ref{Eq:photo_loss}) into the aforementioned UDA model. We can see from Table \ref{tab:ablation} that applying both $L_{RTC}$ (DA-RTC v.s. UDA) and $L_{STC}$ (TCDA v.s. DA-RTC) would yield remarkable improvements over baselines. Qualitative comparisons will be provided in the supplementary material.

\begin{figure}
\begin{center}
\includegraphics[width=0.9\linewidth]{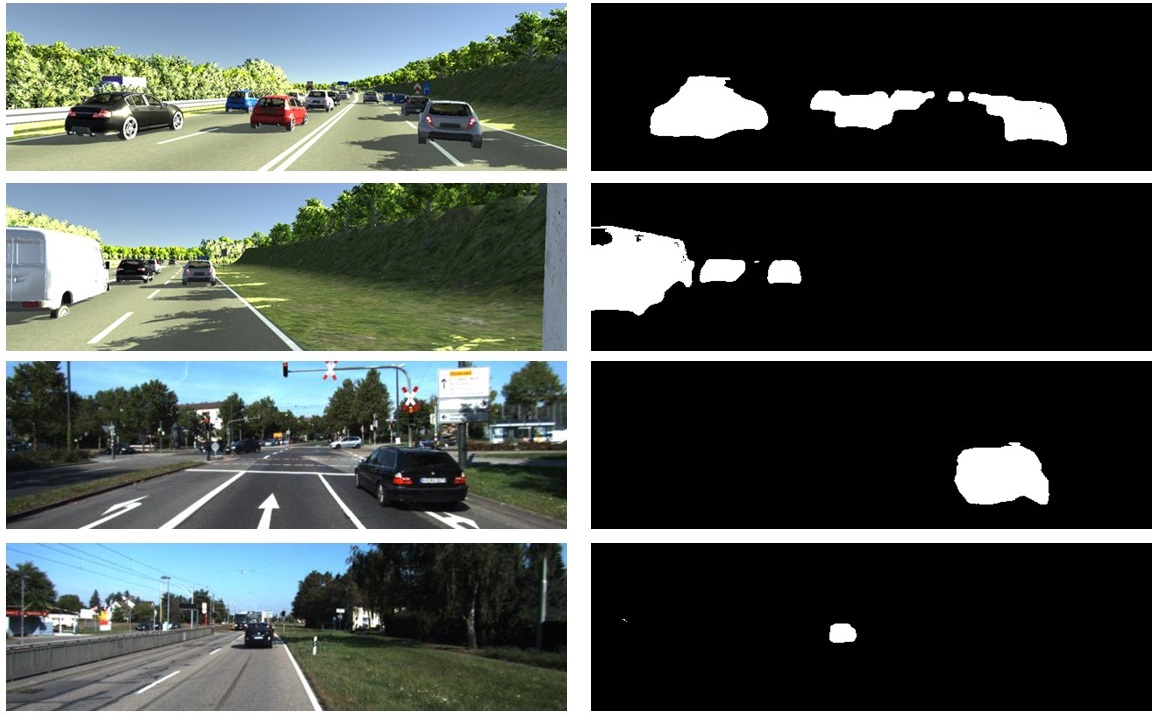}
\end{center}
   \caption{Moving mask predictions on vKITTI and KITTI dataset. The first and second images on the left column are samples from synthetic domain (vKITTI). The third and fourth images on the left column are samples from real domain (KITTI). The corresponding moving masks are shown in the right column.}
\label{fig:short}
\end{figure}

\subsection{Moving Mask}
\begin{figure}
\begin{center}
\includegraphics[width=1\linewidth]{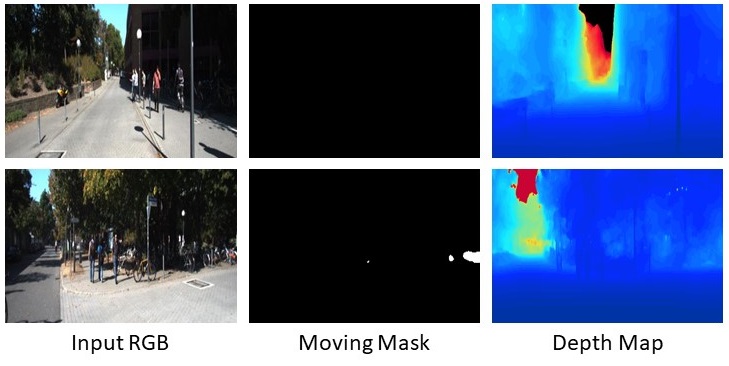}
\end{center}
   \caption{Failure cases. Our model failed on the scene with lots of pedestrians, because there is no pedestrian appearing in the vKITTI \cite{gaidon2016virtual} dataset.}
\label{fig:fail_make3d}
\end{figure}

The camera pose  modular and photometric loss are sensitive to dynamic scenes where objects show irregular movements. Thus, we propose to integrate moving detection to revise the photometric loss, expecting to improve both the camera pose estimation network and the translation network. The moving objects of vKITTI and KITTI are not diverse, but commonly some cars with regular moving trends. However, training TCDA with the guided moving information still improves the performance as shown in Table \ref{tab:ablation} (Sensitive-TCDA v.s. TCDA). Thereby, we believe that the moving mask detection component would show significance in complex scenes with unconstrained moving objects.  
The failure cases in Figure \ref{fig:fail_make3d} also support our analysis. In specific, vKITTI (source domain) does not contain scenes with pedestrians, which degrading the performance of the moving object detection network in recognizing moving pedestrians in target domain. As a result, the depth map misses the person details in some regions. 

\section{Conclusion}
In the paper, we have proposed a temporally-consistent domain adaptation (TCDA) approach to deeply explore the labeled synthetic videos and monocular real videos for monocular depth estimation. We demonstrated that the temporal geometry consistency constraints in both synthetic and real videos play an important role in improving the performance of domain adaptation by improving the quality of translated images as well as the overall depth prediction accuracy. Furthermore, given that the temporal consistency in real videos guided by camera pose and depth is sensitive to the moving objects in the scene, we further proposed a moving mask prediction network trained using synthetic data, which can mask out the moving pixels and thus remove the outlier points in the temporal consistency in real videos. Finally, we proposed to train a camera pose prediction network from synthetic data with camera pose ground truth, which can provide better initialization for estimating the temporal consistency. The deep exploration of synthetic data significantly boosts the effectiveness of domain adaptation and the final depth prediction performance. 

% addressing the issue that camera pose estimation is not robust to dynamic scenes where objects show irregular movements, we propose to 
%photometric error drived depth estimation is sensa
%camera pose guided photometric loss is sensitive to dynamic scenes with unconstrained moving objects

%we integrated the DDVO refined camera pose estimation 
%develop a moving mask guided 

{\small
\bibliographystyle{ieee_fullname}
\bibliography{egbib}
}

\end{document}